%% file: main.tex
\documentclass[10pt,twocolumn,letterpaper]{article}

\usepackage{iccv}
\usepackage{times}
\usepackage{epsfig}
\usepackage{multirow}
\usepackage{graphicx}
\usepackage{amsmath}
\usepackage{amssymb}
\usepackage{amsthm}
\usepackage{bm}
\usepackage{dutchcal}

\usepackage{booktabs}

\usepackage{multirow}
\usepackage{enumitem}
\usepackage{url}            

\usepackage{makecell}

\usepackage{wrapfig}

\usepackage{setspace}
\usepackage{color}

\iccvfinalcopy 

\def\iccvPaperID{6025} 

\ificcvfinal\fi

\begin{document}

\title{ShapeConv: Shape-aware Convolutional Layer \\
for Indoor RGB-D Semantic Segmentation}

\author{
Jinming Cao$^1$\quad
Hanchao Leng$^1$\quad
Dani Lischinski$^2$\quad
Danny Cohen-Or$^3$\quad
Changhe Tu$^1$\thanks{Corresponding Author}\quad
Yangyan Li$^{4*}$
\\

$^1$Shandong University, China\qquad
$^2$The Hebrew University of Jerusalem, Israel\\
$^3$Tel Aviv University, Israel\qquad
$^4$Alibaba Group, China\\

{\tt\small \{jinming.ccao, hanchao.leng, danix3d, cohenor,  changhe.tu, yangyan.lee\}@gmail.com}

}


\maketitle
\ificcvfinal\thispagestyle{empty}\fi

\input{01_abs}
\input{02_intro}

\input{03_rw}

\input{04_method}

\input{05_ex}

\input{06_con}

\noindent
\textbf{Acknowledgments.} This work is supported by the National Key Research and Development Program of China grant No.2017YFB1002603, the National Science Foundation of China General Program grant No.61772317, 61772318 and 62072284, “Qilu” Young Talent Program of Shandong University, and the Research Intern Program of Alibaba Group.

\bibliographystyle{ieee_fullname}
\bibliography{main}

\end{document}

%% file: 01_abs.tex
\begin{abstract}

RGB-D semantic segmentation has attracted increasing attention over the past few years. 
Existing methods mostly employ homogeneous convolution operators to consume the RGB and depth features, ignoring their intrinsic differences. In fact, the RGB values capture the photometric appearance properties in the projected image space, while the depth feature encodes both the \emph{shape} of a local geometry as well as the \emph{base} (whereabout) of it in a larger context. 
Compared with the \emph{base}, the \emph{shape} probably is more inherent and has a stronger connection to the semantics, and thus is more critical for segmentation accuracy. 
Inspired by this observation, we introduce a Shape-aware Convolutional layer (ShapeConv) for processing the depth feature, where the depth feature is firstly decomposed into a shape-component and a base-component, next two learnable weights are introduced to cooperate with them independently, and finally a convolution is applied on the re-weighted combination of these two components.
ShapeConv is model-agnostic and can be easily integrated into most CNNs to replace vanilla convolutional layers for semantic segmentation.
Extensive experiments on three challenging indoor RGB-D semantic segmentation benchmarks, i.e., NYU-Dv2(-13,-40), SUN RGB-D, and SID, demonstrate the effectiveness of our ShapeConv when employing it over five popular architectures.
Moreover, the performance of CNNs with ShapeConv is boosted without introducing any computation and memory increase in the inference phase. The reason is that the learnt weights for balancing the importance between the shape and base components in ShapeConv become constants in the inference phase, and thus can be fused into the following convolution, resulting in a network that is identical to one with vanilla convolutional layers.

\end{abstract}

%% file: 02_intro.tex
\section{Introduction}
\label{sec:intro}
\vspace{-0.2cm}
With the widespread use of depth sensors (such as Microsoft Kinect~\cite{zhang2012microsoft}), the availability of RGB-D data has boosted the advancement of RGB-D semantic segmentation, which contributes to an indispensable task in the computer vision community. Thanks to the flourishing of Convolutional Neural Networks (CNNs), recent studies mostly resort to CNNs for tackling this problem. Convolutional layers, deemed as the core building blocks of CNNs, are accordingly the key elements in RGB-D semantic segmentation models~\cite{cheng2017locality,hu2019acnet,li2020attention,lin2017cascaded,park2017rdfnet}. 

\begin{figure}[t!]
	\centering
	\includegraphics[width=0.45\textwidth]{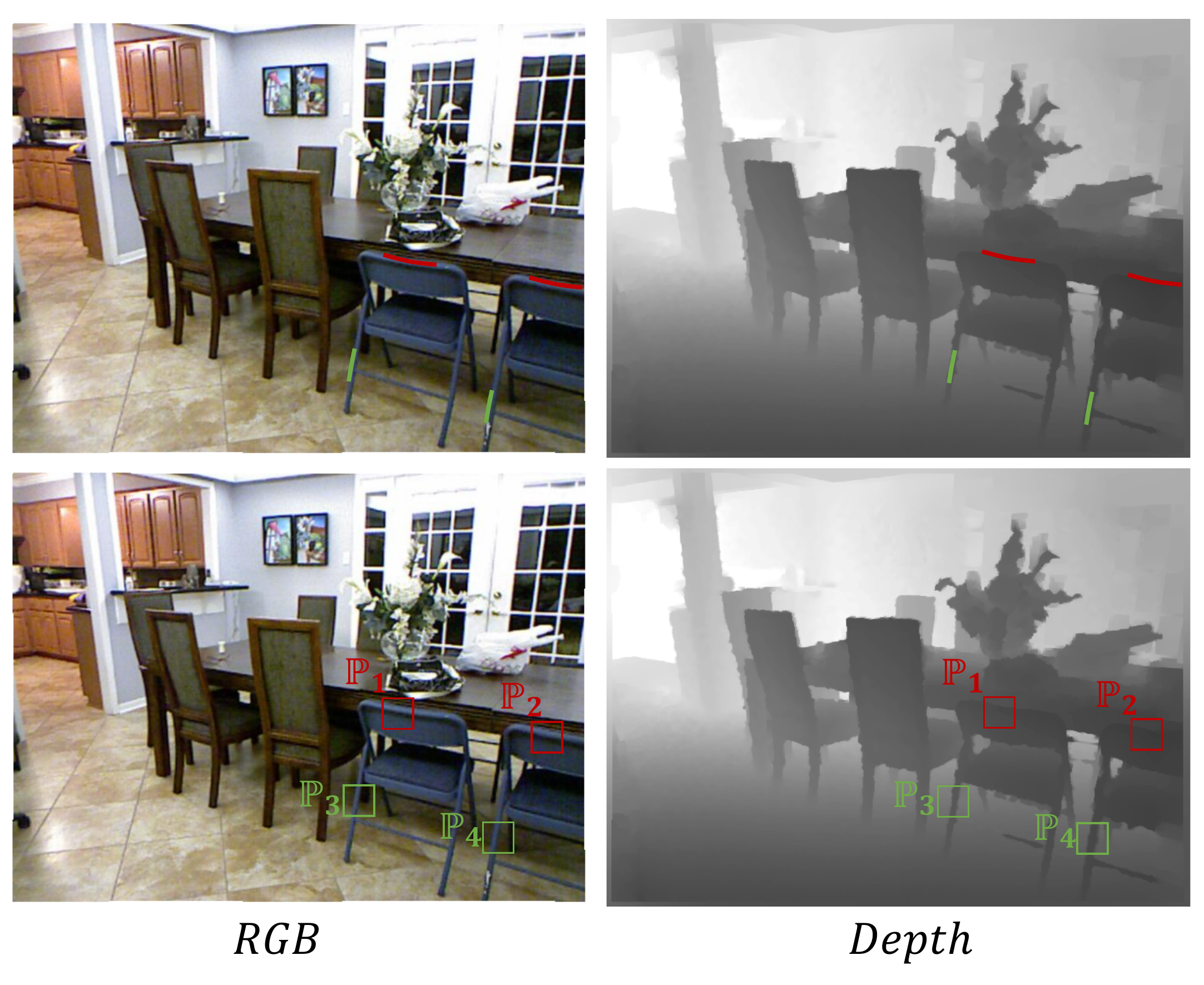}
	\vspace{-0.03\linewidth}
	\caption{Visual demonstration of why the \emph{shape} of an RGB-D image matters. Regarding the images on the top, lines with the same color share a same \emph{shape}, yet with different \emph{base}. The corresponding patches are shown on the bottom.}
	\label{fig:two-patch}
\end{figure}

However, RGB and depth information are inherently different from each other. In particular, RGB values capture the photometric appearance properties in the projected image space, while the depth feature encodes both the \emph{shape} of a local geometry as well as the \emph{base} (whereabout) of it in a larger context. As a result, the convolution operator that is widely adopted for consuming RGB data might not be the optimal for processing the depth data.
Taking Figure~\ref{fig:two-patch} as an example, we would expect the corresponding patches of the same chairs to have the same features, as they share the same \emph{shape}. The shape is a more inherent property of the underlying object and has stronger connection to the semantics. We would expect to achieve shape invariance in the learning process. When a vanilla convolution operator is applied on these corresponding patches, the resulting features are different due to the differences in their \emph{base} component, hindering the learning from achieving shape invariance. On the other hand, the \emph{base} components cannot be simply discarded for pursuing the shape invariance in the current layer, as they form the \emph{shape} in a followup layer with a larger context.
%
%

To address these problems, we propose a \textbf{Shape}-aware \textbf{Conv}lutional layer (ShapeConv), to learn the adaptive balance between the importance of \emph{shape} and \emph{base} information, giving the network the chance to focus more on the \emph{shape} information whenever necessary for benefiting the RGB-D semantic segmentation task.
We firstly decompose a patch\footnote{The operation unit of input features for the convolutional layer, whose spatial size is the same as the convolution kernel.} into two separate components, i.e., a \emph{base-component} and a \emph{shape-component}. The mean of patch values depicts the whereabout of the patch in a larger context, thus constitutes the base component, while the residual is the relative changes in the patch, which depicts the shape of the underlying geometry, thus constitutes to the shape component. Specifically, for an input patch (such as $\mathbb{P}_1$ in Figure~\ref{fig:two-patch}), the \emph{base} describes where the patch is, i.e., the distance from the observation point; while the \emph{shape} expresses what the patch is, e.g., a chair corner.
We then employ two operations, namely, base-product and shape-product, to respectively process these two components with two learnable weights, i.e., base-kernel and shape-kernel. The output from these two is then combined in an addition manner to form a shape-aware patch, which is further convolved with a normal convolutional kernel. In contrast to the original patch, the shape-aware one is capable of adaptively learning the shape characteristic with the shape-kernel, and the base-kernel serves to balance the contributions of the \emph{shape} and the \emph{base} for the final prediction.

In addition, since the base-kernel and shape-kernel become constants in the inference phase, we can fuse them into the following convolution kernel, resulting in a network that is identical to the one with vanilla convolutional layers. The proposed ShapeConv can be easily plugged into most CNNs as a replacement of the vanilla convolution in semantic segmentation without introducing any computation and memory increase in the inference phase. This simple replacement transforms CNNs designed for RGB data into ones better suited for consuming RGB-D data.

To validate the effectiveness of the proposed method, we conduct extensive experiments on three challenging RGB-D indoor semantic segmentation benchmarks: NYU-Dv2~\cite{silberman2012indoor}(-13,-40), SUN RGBD~\cite{song2015sun}, and SID~\cite{armeni2017joint}. We apply our ShapeConv to five popular semantic segmentation architectures and can observe promising performance improvements compared with baseline models. We found that ShapeConv can significantly improve the segmentation accuracy around the object boundaries (see Figure~\ref{fig:trimap}), which demonstrates the effective leveraging of the depth information\footnote{Our code is released through {https://github.com/hanchaoleng/ShapeConv}.}.

%% file: 03_rw.tex
\section{Related Work}
\vspace{-0.2cm}
CNNs have been widely used for semantic segmentation on RGB images~\cite{chen2017rethinking,chen2018encoder,long2015fully,lin2017feature,ronneberger2015u,zhao2017pyramid}. In general, existing segmentation architectures usually involve two stages: the backbone and the segmentation stage. The former stage is leveraged to extract features from RGB images, wherein popular models are ResNet~\cite{he2016deep}, ResNeXt~\cite{xie2017aggregated} which are pre-trained on the ImageNet dataset~\cite{russakovsky2015imagenet}. The latter stage aims to generate predictions based on the extracted features. Methods in this stage include Upsample~\cite{long2015fully}, PPM~\cite{zhao2017pyramid} and ASPP~\cite{chen2017rethinking,chen2018encoder}, etc. It is worth noting that both stages adopt the convolutional layers as the core building blocks.

As RGB semantic segmentation has been extensively studied in literature, a straightforward solution for RGB-D semantic segmentation is to adapt the well-developed architectures from the ones designed for RGB data. However, implementing such a idea is non-trivial due to the asymmetric modality problem between the RGB and the depth information. To tackle this, researchers have devoted efforts into two directions: designing dedicated architectures for RGB-D data~\cite{cheng2017locality,fooladgar2019multi,hu2019acnet,li2020attention,lin2017cascaded,park2017rdfnet,wang2020deep}, and presenting novel layers to enhance or replace the convolutional layers in RGB semantic segmentation~\cite{chen2021spatial,wang2018depth,xing2020malleable}. Our method falls into the second category.

Methods in the first category propose to feed RGB and depth channels to two parallel CNNs streams, where the output features are fused with specific strategies. For example,~\cite{cheng2017locality} presents a gate-fusion method,~\cite{fooladgar2019multi,hu2019acnet,park2017rdfnet} fuse the features in multi-levels of the backbone stages.
Nevertheless, these methods mostly leverage separate networks to consume RGB and depth features, they are yet faced with two limitations: 1) it is hard to decide when is the best stage for the fusion to happen; and 2) the two-stream or multi-level way often results in large increase of computation.

In contrast, methods along the second direction target at designing novel layers based on the geometric characteristics of RGB-D data, which are more flexible and time-efficient. For instance, Wang \etal~\cite{wang2018depth} proposed the depth-aware convolution to weight pixels based on a hand-crafted Gaussian function by leveraging the depth similarity between pixels.~\cite{xing2020malleable} presents a novel operator called malleable 2.5D convolution, to learn the receptive field along the depth-axis.~\cite{chen2021spatial} devises a S-Conv to infer the sampling offset of the convolution kernel guided by the 3D spatial information, enabling the convolutional layer to adjust the receptive field and geometric transformations. ShapeConv proposed a novel view of the content in each patch and a mechanism to leverage them adaptively with learnt weights. Moreover, ShapeConv can be converted into vanilla convolution in the inference phase, resulting in ZERO increase of memory and computation compared with the models with vanilla convolution.

%% file: 04_method.tex
\vspace{-0.2cm}
\section{Method}
\vspace{-0.2cm}
In this section, we first provide the basic formulation of the Shape-aware convolutional layer (ShapeConv) for RGB-D data, followed by its application in the training and inference phase. We end this section with the method architectures. 
\vspace{-0.1cm}
\subsection{ShapeConv for RGB-D Data}
\label{sec:method-1}
\vspace{-0.1cm}

\paragraph{Method Intuition.} Given an input patch $\mathbb{P} \in R^{K_h \times K_w \times C_{in}}$, $K_h$ and $K_w$ are the spatial dimensions of the kernel; $C_{in}$ represents the channel numbers in the input feature map, the output features from the vanilla convolution layer are obtained by,
\begin{equation}
\label{eq:conv}
    \mathbb{F} = Conv(\mathbb{K}, \mathbb{P}),
\end{equation}
where $\mathbb{K} \in R^{K_h \times K_w \times C_{in} \times C_{out}}$ denotes the learnable weights of kernels in a convolutional layer (The bias terms are not included for simplicity.); $C_{out}$ represents the channel numbers in the output feature map. Each element of $\mathbb{F} \in R^{C_{out}}$ is calculated as,
\begin{equation*}
   \mathbb{F}_{c_{out}} = \sum_{i}^{K_h \times K_w \times C_{in}} (\mathbb{K}_{i,c_{out}} \times \mathbb{P}_i).
\label{eq:conv-f}
\end{equation*}

It can be easily recognized that $\mathbb{F}$ usually changes with respect to different values of $\mathbb{P}$. Take the two patches in the Figure~\ref{fig:two-patch}, $\mathbb{P}_1$ and $\mathbb{P}_2$, as an example. The corresponding output features, $\mathbb{F}_1$ and $\mathbb{F}_2$ from the vanilla convolution layer are learned by: $\mathbb{F}_1 = Conv(\mathbb{K}, \mathbb{P}_1)$, $\mathbb{F}_2 = Conv(\mathbb{K}, \mathbb{P}_2)$. Since $\mathbb{P}_1$ and $\mathbb{P}_2$ are not identical (different distances from the observation points), accordingly, their features are usually different, and this may lead to distinct prediction results.

Nevertheless, $\mathbb{P}_1$ and $\mathbb{P}_2$, corresponding to the red regions in Figure~\ref{fig:two-patch}, actually belong to the same class - chair. And vanilla convolutional layers cannot well handle such situations. In fact, there exists some invariants of these two patches, namely, the \emph{shape}. It refers to the relative depth correlation under local features, which is however, unexpectedly ignored by the existing methods. In view of this, we propose to fill this gap via effectively modeling the \emph{shape} for RGB-D semantic segmentation.

\vspace{-0.4cm}
\paragraph{ShapeConv Formulation.} Based on the aforementioned analysis, in this paper, we offer to decompose an input patch into two components: a base-component $\mathbb{P}_B$ describing where the patch is, and a shape-component $\mathbb{P}_S$ expressing what the patch is. Therefore, we refer the mean\footnote{As the depth values are obtained from a fixed observation point, we notice that the rotational transformations cannot be addressed due to the angle of view limitation. As a result, we focus more on the translational transformations in this paper.} of patch values to be $\mathbb{P}_B$, and its relative values to be as $\mathbb{P}_S$:
\begin{equation*}
   \begin{aligned}
    & \mathbb{P}_B = m(\mathbb{P}), \\
    & \mathbb{P}_S = \mathbb{P} - m(\mathbb{P}),
   \end{aligned}
\end{equation*}
where $m(\mathbb{P})$ is the mean function on $\mathbb{P}$ (over the $K_h \times K_w$ dimensions), and $\mathbb{P}_B \in R^{1 \times 1 \times C_{in}}$, and $\mathbb{P}_S \in R^{K_h \times K_w \times C_{in}}$.

Note that directly convolved $\mathbb{P}_S$ with $\mathbb{K}$ in Equation~\ref{eq:conv} is sub-optimal, as the values from $\mathbb{P}_B$ contributes the class discrimination across patches. Thus, our ShapeConv instead leverages two learnable weights, $\mathbb{W}_B \in R^{1}$ and $\mathbb{W}_S \in R^{K_h \times K_w \times K_h \times K_w \times C_{in}}$ , to separately consume the above two components. The outputted features are then combined in an element-wise addition manner, which forms a new shape-aware patch with the same size as the original one $\mathbb{P}$. The formulation of ShapeConv is given as,

\begin{equation}
\label{eq:shapeconv-patch}
\begin{aligned}
 \mathbb{F} &= ShapeConv(\mathbb{K}, \mathbb{W}_B, \mathbb{W}_S, \mathbb{P})\\
              &= Conv(\mathbb{K}, \mathbb{W}_B \diamond \mathbb{P}_B + \mathbb{W}_S \ast \mathbb{P}_S)\\
              &= Conv(\mathbb{K}, \textbf{P}_\textbf{B} + \textbf{P}_\textbf{S})\\
              &= Conv((\mathbb{K}, \textbf{P}_\textbf{BS}),
\end{aligned}
\end{equation}
where $\diamond$ and $\ast$ denote the base-product and shape-product operator, respectively, which are defined as,
\begin{equation}
\begin{cases}
   \textbf{P}_\textbf{B} = \mathbb{W}_B \diamond \mathbb{P}_B \\
   \textbf{P}_{\textbf{B}_{1,1,c_{in}}} = \mathbb{W}_B \times \mathbb{P}_{B_{1,1,c_{in}}},
\end{cases}
\label{eq:base-product-P}
\end{equation}
\begin{equation}
\begin{cases}
    \textbf{P}_\textbf{S} = \mathbb{W}_S \ast \mathbb{P}_S \\
   \textbf{P}_{\textbf{S}_{{k_h},{k_w},{c_{in}}}} = \sum_{i}^{K_h \times K_w} (\mathbb{W}_{S_{i,{k_h},{k_w},{c_{in}}}} \times \mathbb{P}_{S_{i,{c_{in}}}}),
\end{cases}
\label{eq:shape-product-P}
\end{equation}
where $c_{in}$, $k_h$, $k_w$ are the indices of the elements in $C_{in}$, $K_h$, $K_w$ dimensions, respectively.

\begin{figure}[t!]
	\centering
	\includegraphics[width=0.45\textwidth]{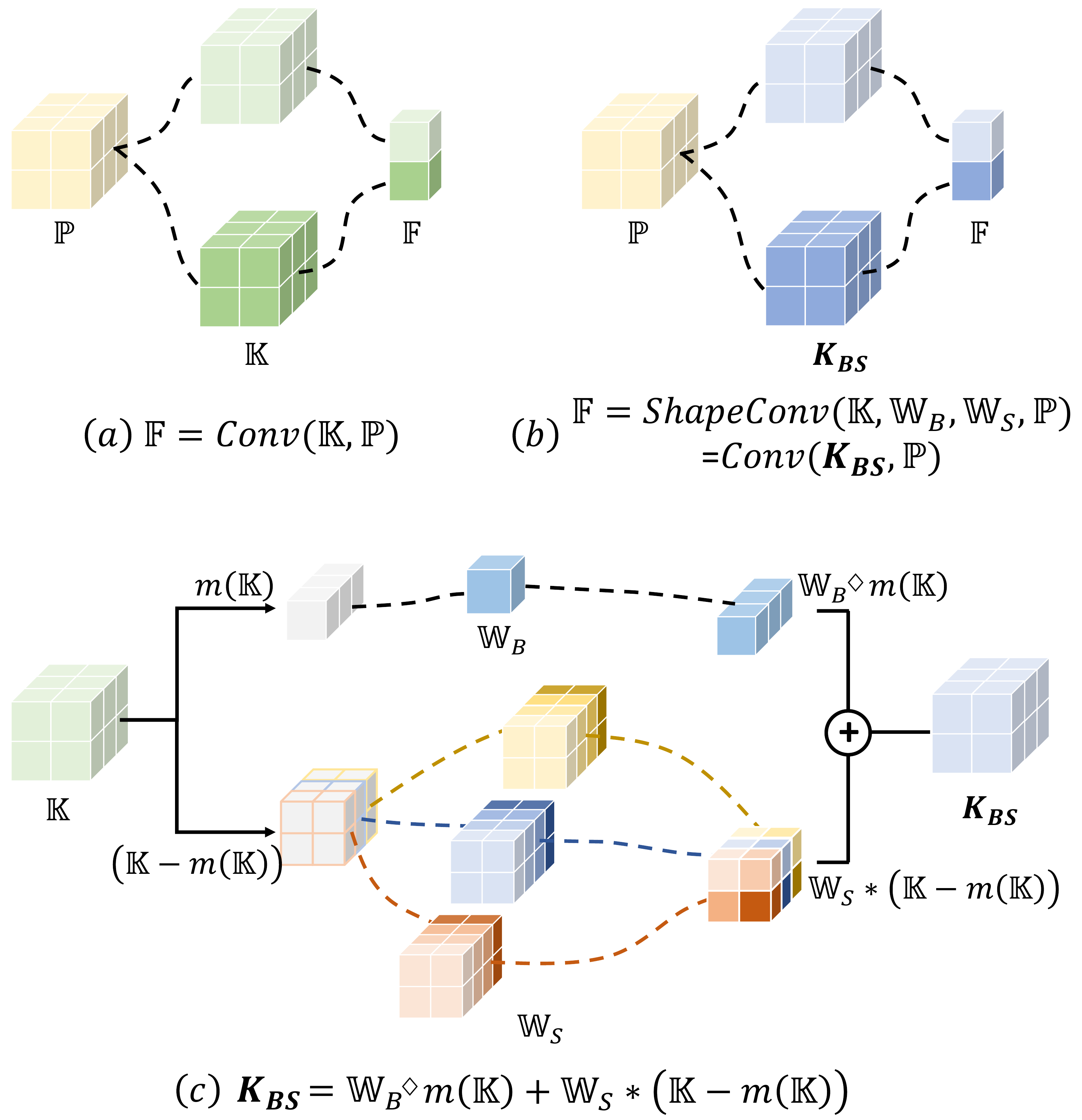}
	\caption{Comparison of vanilla convolution and ShapeConv within a patch $\mathbb{P}$. In this figure, $K_h = K_w = 2$, $C_{in} = 3$, and $C_{out} = 2$, ``+'' denotes element-wise addition. (a) Vanilla convolution with kernel $\mathbb{K}$; (b) ShapeConv with folding the $\mathbb{W}_B$ and $\mathbb{W}_S$ into $\textbf{K}_\textbf{BS}$; (c) The computation of $\textbf{K}_\textbf{BS}$ from $\mathbb{K}$, $\mathbb{W}_B$ and $\mathbb{W}_S$.
	}
	\label{fig:ShapeConv}
\end{figure}

We reconstruct the shape-aware patch $\textbf{P}_\textbf{BS}$ from the addition of $\textbf{P}_\textbf{B}$ and $\textbf{P}_\textbf{S}$, and $\textbf{P}_\textbf{BS} \in R^{K_h \times K_w \times C_{in}}$, which enables it to be smoothly convolved by the kernel $\mathbb{K}$ of vanilla convolutional layer.
Nevertheless, the $\textbf{P}_\textbf{BS}$ is equipped with the important shape information which is learned by the two additional weights, making the convolutional layer to focus on the situations when merely using depth values fails.

\vspace{-0.1cm}
\subsection{ShapeConv in Training and Inference}
\vspace{-0.1cm}
\paragraph{Training phase.} The proposed ShapeConv in Section~\ref{sec:method-1} can effective leverage the \emph{shape} information of patches. However, replacing vanilla convolutional layer with ShapeConv in CNNs introduces more computational cost due to the two \emph{product} operation in Equation~\ref{eq:base-product-P} and ~\ref{eq:shape-product-P}. To tackle this problem, we propose to shift these two operations from patches to kernels,
\begin{equation*}
\begin{aligned}
\begin{cases}
    \textbf{K}_\textbf{B} = \mathbb{W}_B \diamond \mathbb{K}_B \\
   \textbf{K}_{\textbf{B}_{1,1,c_{in},c_{out}}} = \mathbb{W}_B \times \mathbb{K}_{B_{1,1,c_{in},c_{out}}},
\end{cases}
\label{eq:base-product-K}
\end{aligned}
\end{equation*}
\begin{equation*}
\begin{aligned}
\begin{cases}
     \textbf{K}_\textbf{S} = \mathbb{W}_S \ast \mathbb{K}_S \\
   \textbf{K}_{\textbf{S}_{{k_h},{k_w},{c_{in}},{c_{out}}}} = \sum_{i}^{K_h \times K_w} (\mathbb{W}_{S_{i,{k_h},{k_w},{c_{in}}}} \times \mathbb{K}_{S_{i,{c_{in}},{c_{out}}}}),
\end{cases}
\label{eq:shape-product-K}
\end{aligned}
\end{equation*}
where $\mathbb{K}_B \in R^{1 \times 1 \times C_{in} \times C_{out}}$ and $\mathbb{K}_S \in R^{K_h \times K_w \times C_{in} \times C_{out}}$ denote the base-component of kernels and shape-component, respectively, and $\mathbb{K} = \mathbb{K}_B + \mathbb{K}_S$. 

We therefore re-formalize ShapeConv the Equation~\ref{eq:shapeconv-patch} to following:
\begin{equation}
\label{eq:shapeconv-kernel}
\begin{aligned}
 \mathbb{F} &= ShapeConv(\mathbb{K}, \mathbb{W}_B, \mathbb{W}_S, \mathbb{P})\\
              &= Conv(\mathbb{W}_B \diamond m(\mathbb{K})+ \mathbb{W}_S \ast (\mathbb{K} - m(\mathbb{K})), \mathbb{P})\\
              &= Conv(\mathbb{W}_B \diamond \mathbb{K}_B+ \mathbb{W}_S \ast \mathbb{K}_S, \mathbb{P})\\
              &= Conv(\textbf{K}_\textbf{B} + \textbf{K}_\textbf{S}, \mathbb{P})\\
              &= Conv(\textbf{K}_\textbf{BS}, \mathbb{P}),
\end{aligned}
\end{equation}
where $m(\mathbb{K})$ is the mean function on $\mathbb{K}$ (over the $K_h \times K_w$ dimensions). And we require $\textbf{K}_\textbf{BS} = \textbf{K}_\textbf{B} + \textbf{K}_\textbf{S}$, $\textbf{K}_\textbf{BS} \in R^{K_h \times K_w \times C_{in} \times C_{out}}$.

In fact, the two formulations of ShpeConv, i.e., Equation~\ref{eq:shapeconv-patch} and Equation~\ref{eq:shapeconv-kernel} are mathematically equivalent, i.e.,

\begin{equation}
\begin{aligned}
\mathbb{F} &= ShapeConv(\mathbb{K}, \mathbb{W}_B, \mathbb{W}_S, \mathbb{P})\\
              &= Conv(\mathbb{K}, \textbf{P}_\textbf{BS})\\
              &= Conv(\textbf{K}_\textbf{BS}, \mathbb{P}),
\end{aligned}
\end{equation}

\begin{equation}
\begin{aligned}
     \mathbb{F}_{c_{out}} &= \sum_{i}^{K_h \times K_w \times C_{in}} (\mathbb{K}_{i,{c_{out}}} \times \textbf{P}_{\textbf{BS}_{i}})\\
     &= \sum_{i}^{K_h \times K_w \times C_{in}} (\textbf{K}_{\textbf{BS}_{i,{c_{out}}}} \times \mathbb{P}_{i}),
\end{aligned}
\end{equation}
please refer to the Supp. for detailed proof. In this way, we utilize the ShapeConv in Equation~\ref{eq:shapeconv-kernel} in our implementation as illustrated in Figure~\ref{fig:ShapeConv}(b) and (c).

\vspace{-0.5cm}
\paragraph{Inference phase.} 
During inference, since the two additional weights i.e. $\mathbb{W}_B$ and $\mathbb{W}_S$, become constants, we can fuse them into $\textbf{K}_\textbf{BS}$ as shown in Figure~\ref{fig:ShapeConv}(c) with $\textbf{K}_\textbf{BS} = \mathbb{W}_B \diamond \mathbb{K}_B+ \mathbb{W}_S \ast \mathbb{K}_S$. 
And $\textbf{K}_\textbf{BS}$ shares the same tensor size with $\mathbb{K}$ in Equation~\ref{eq:conv}, thus, our ShapeConv is actually the same as the vanilla convolutional layer in Figure~\ref{fig:ShapeConv}(a). In other words, when replacing vanilla convolution with ShapeConv, there would introduce zero additional inference time.

\begin{figure*}[t!]
	\centering
	\centering
	\includegraphics[width=\textwidth]{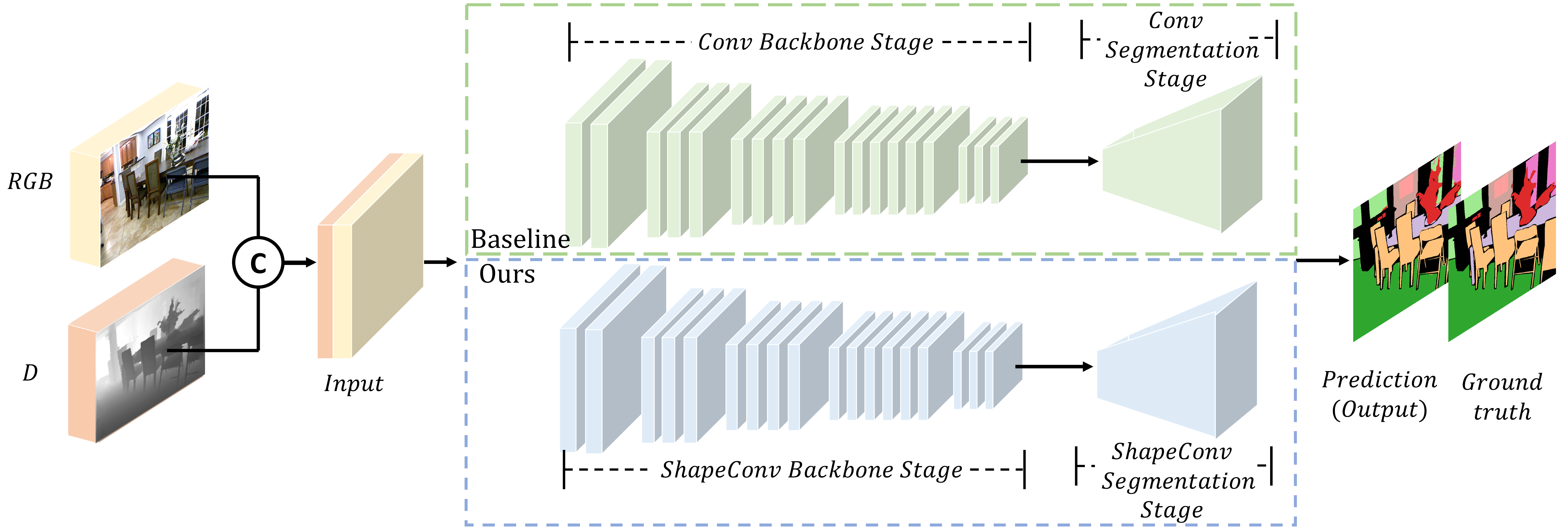}
	\vspace{-0.03\linewidth}
	\caption{The overall semantic segmentation network architecture. In this figure, yellow and orange cube denote the RGB and D inputs; ``C'' denotes channel-wise concatenation; Green and blue boxes denote architectures consisting of vanilla convolutional layers and ShapeConv layers, respectively.}
	\label{fig:Architecture}
\end{figure*}
\vspace{-0.1cm}
\subsection{ShapeConv-enhanced Network Architecture}
\vspace{-0.1cm}
Different from devising specially dedicated architectures for RGB-D segmentation~\cite{park2017rdfnet,qi20173d,lin2017cascaded}, the proposed ShapeConv is a more generalized approach that can be easily plugged into most CNNs as a replacement for the vanilla convolution in semantic segmentation, which is then transformed for adapting the RGB-D data.

Figure~\ref{fig:Architecture} depicts an example of the overall method architecture. In order to leverage the advanced backbones in semantic segmentation, we firstly require to convert the input features from RGB images to RGB-D data via the concatenation of the RGB and D information. In practice, D can be depth values~\cite{hazirbas2016fusenet,ma2017multi} or HHA\footnote{Horizontal disparity, Height above ground and normal Angle to the vertical axis.} images~\cite{gupta2014learning,long2015fully,li2016lstm,cheng2017locality}. We then replace the vanilla convolution layer with the ShapeConv in both the backbone and segmentation stages. It is worth noting that, $\mathbb{W}_B$ is initialized to one, $\mathbb{W}_S$ can be viewed as $C_{in}$ square $(K_h \times K_w) \times (K_h \times K_w)$ matrices, which are initialized to the identity matrix. In this way, ShapeConv is equivalent to the vanilla convolution at the beginning of training since $\textbf{K}_\textbf{BS}=\mathbb{K}$. This initialization approach offers two advantages: 1) It makes the ShapeConv-enhanced networks do not interfere with the RGB data, i.e., the RGB features are processed in the same way as before. 2) It facilitates ShapeConv to reuse the parameters from pre-trained models.

Thus, with this approach, future advances in RGB semantic segmentation architectures can be easily transferred to consuming the RGB-D data, greatly reducing the effort that would otherwise be spent on designing dedicated networks for RGB-D semantic segmentation. We have shown the results of building RGB-D segmentation networks with this style using several popular architectures~\cite{chen2017rethinking,chen2018encoder,lin2017feature,ronneberger2015u,zhao2017pyramid} in Sec~\ref{sec:diff ar}.

%% file: 05_ex.tex
\vspace{-0.3cm}
\section{Experiments}
\vspace{-0.2cm}
\paragraph{Datasets and metrics.}
Among the existing RGB-D segmentation problems, the indoor semantic segmentation is rather challenging, as the objects are often complex and with severe occlusions~\cite{chen2021spatial}. Thus, in order to validate the effectiveness of the proposed method, we conducted experiments on three indoor RGB-D benchmarks: NYU-Depth-V2 (NYUDv2-13 and -40)~\cite{silberman2012indoor}, SUN-RGBD~\cite{song2015sun} and Stanford Indoor Dataset (SID)~\cite{armeni2017joint}. NYUDv2 contains 1,449 RGB-D scene images, where 795 images are split for training and 654 images for testing. We adopted two popular settings for this dataset, i.e., 13-class~\cite{silberman2012indoor} and 40-class~\cite{gupta2013perceptual}, where all pixels are labeled with 13 and 40 classes, respectively. SUN-RGBD is composed of 10,355 RGB-D indoor images with 37 categories for each pixel label. We followed the widely used setting in ~\cite{song2015sun} to split the dataset into a training set of 5285 images and a testing set of 5050 images. SID contains 70, 496 RGB-D images with 13 object categories. In particular, areas 1, 2, 3, 4, and 6 used for the training and Area 5 is for testing following~\cite{wang2018depth}.

We reported the results using the same evaluation protocol and metrics as FCN~\cite{long2015fully}, i.e., Pixel Accuracy (\emph{Pixel Acc.}), Mean Accuracy (\emph{Mean Acc.}), Mean Region Intersection Over Union (\emph{Mean IoU}), and Frequency Weighted Intersection Over Union (\emph{f.w. IoU}).

\vspace{-0.6cm}
\paragraph{Comparison protocol.} We adopted several popular architectures with different backbones as our baseline methods to demonstrate the effectiveness and generalization capability of ShapeConv. For all the baseline methods, we only replaced the vanilla convolutional layers with our ShapeConv, without any change to other settings. This guarantees that the obtained performance improvements is due to the application of ShapeConv, but not other factors. 

\begin{table}[h!]
\centering
\caption{
\label{tab:nyu-13}
Performance comparison with baselines on NYUDv2-13 dataset. Deeplabv3+ is the adopted architecture. }
\scalebox{0.8}{
\begin{tabular}{c|c|c|c|c|c}
\hline
Back  &\multirow{2}{*}{Setting} & Pixel  & Mean   & Mean  & f.w. \\
 bone                          &         & Acc.(\%) &Acc.(\%)  &IoU.(\%)  &IoU.(\%) \\
                        \hline
                                   & Baseline  & 80.0 & 72.5      & 60.8   & 67.6      \\ \cline{2-6} 
                                   & Baseline$^{\bigstar}$        & 80.6 & 72.7      & 61.6   & 68.5     \\ \cline{2-6} 
  ResNet                           & Ours & 80.4 & 73.0      & 61.8   & 68.1      \\ \cline{2-6} 
  50~\cite{he2016deep}             & Ours$^{\bigstar}$         & 81.1 & 73.4      & 62.7    & 69.1      \\ \cline{2-6} 
                                   & +  & 0.4  & 0.5       & 1.0     & 0.5      \\ \cline{2-6} 
                                   & +$^{\bigstar}$    & 0.5  & 0.7  & 1.1     & 0.6       \\ \hline
                                   & Baseline  & 80.0 & 73.4      & 61.3    & 67.6      \\ \cline{2-6} 
                                   & Baseline$^{\bigstar}$        & 81.0 & 74.3      & 63.1    & 68.9     \\ \cline{2-6} 
   ResNet                          & Ours & 81.2 & 74.9      & 62.9   & 69.1     \\ \cline{2-6} 
    101~\cite{he2016deep}         & Ours$^{\bigstar}$        & 81.9 & 75.7      & 64.0    & 70.1      \\ \cline{2-6} 
                                   & +   & 1.2  & 1.5      & 1.6     & 1.5      \\ \cline{2-6} 
                                   & +$^{\bigstar}$   & 0.9  & 1.4      & 0.9      & 1.2       \\ \hline
                                   & Baseline  & 81.8& 73.9     & 63.2   & 70.1      \\ \cline{2-6} 
                                   & Baseline$^{\bigstar}$       & 82.2 & 74.4      & 63.7   & 70.6     \\ \cline{2-6} 
  ResNext                       & Ours & 82.6 & 75.7      & 65.1    & 71.2      \\ \cline{2-6} 
 101\_32x8d                     & Ours$^{\bigstar}$        & 82.9 & 76.0      & 65.6    & 71.6      \\ \cline{2-6} 
  ~\cite{xie2017aggregated}     & +         & 0.8  & 1.8       & 1.9   & 1.1       \\ \cline{2-6} 
                                   & +$^{\bigstar}$  & 0.7  & 1.6      & 1.9   & 1.0       \\ \hline
\end{tabular}
}

\end{table}  

\vspace{-0.6cm}
\paragraph{Implementation Details.} We used the ResNet~\cite{he2016deep} and ResNeXt~\cite{xie2017aggregated} initialized with the pre-trained model on ImageNet~\cite{russakovsky2015imagenet} in the backbone stage. If not otherwise noted, the inputs of both the baseline and ours are the concatenation of RGB and HHA images. We adopted both single-scale and multi-scale testing strategies during inference. For the latter one, left-right flipped images and five scales are exploited: [0.5, 0.75, 1.0, 1.25, 1.5, 1.75]. $^{\bigstar}$ in tables of this section denotes the multi-scale strategy. Note that, no post-processing tricks like CRF~\cite{chen2017deeplab} is used in our experiments.

\begin{figure*}[t!]
	\centering
	\centering
	\includegraphics[width=\textwidth]{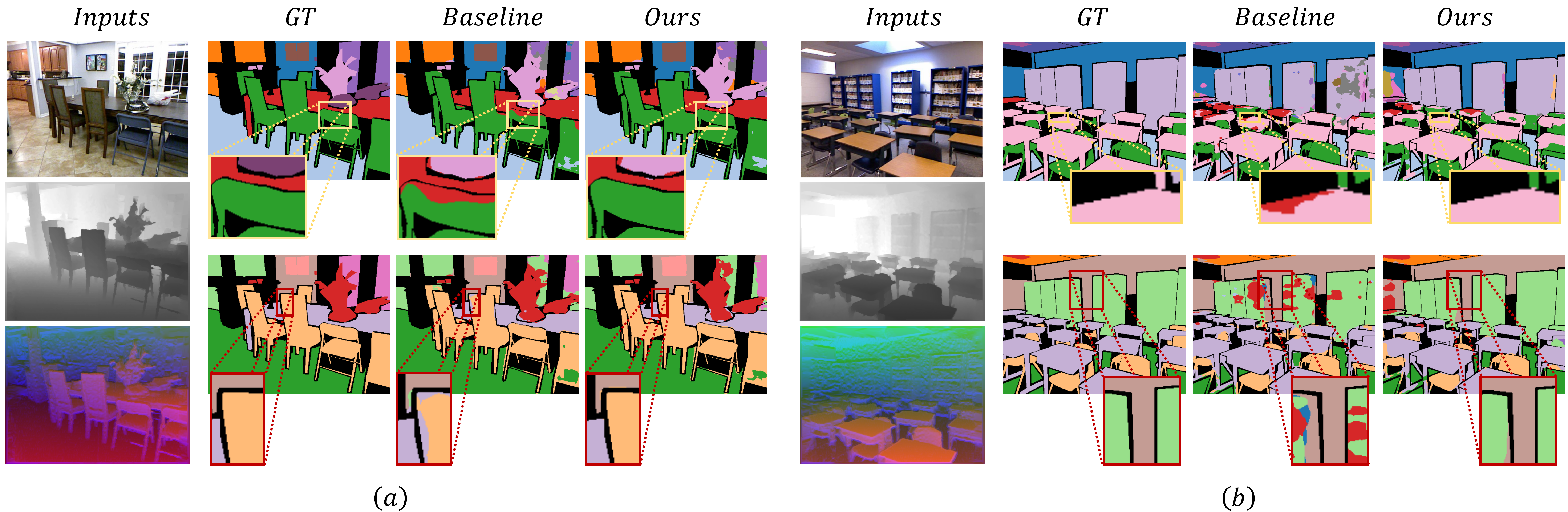}
	\caption{Visualization results from NYUDv2 dataset. Input column denotes RGB, Depth, HHA images from top to bottom; the black regions in the GT, Baseline and Ours indicate the ignored category. The upper and lower cases are from NYUDv2-40 and NYUDv2-13, respectively.}
	\label{fig:vis-seg}
\end{figure*}

\begin{table}[h!]
\centering

\caption{
\label{tab:nyu-40}
Performance comparison with baselines on NYUDv2-40 dataset. Deeplabv3+ is the adopted architecture.}
\scalebox{0.8}{
\begin{tabular}{c|c|c|c|c|c}
\hline
Back  &\multirow{2}{*}{Setting} & Pixel  & Mean   & Mean  & f.w. \\
bone                           &         & Acc.(\%) &Acc.(\%)  &IoU.(\%)  &IoU.(\%) \\
                        \hline
                                   & Baseline  & 73.1 & 57.7      & 45.6   & 59.2     \\ \cline{2-6} 
                                   & Baseline$^{\bigstar}$        & 74.2 & 59.0    & 47.1   & 60.2    \\ \cline{2-6} 
  ResNet                         & Ours & 74.1 & 59.1     & 47.3  & 60.5     \\ \cline{2-6} 
  50~\cite{he2016deep}           & Ours$^{\bigstar}$    & 75.0 & 60.4     & 48.8  & 61.4     \\ \cline{2-6} 
                                   & +       & 1.0  & 1.4     & 1.7   & 1.3      \\ \cline{2-6} 
                                   & +$^{\bigstar}$    & 0.8 & 1.4      & 1.7    & 1.2    \\ \hline
                                   & Baseline  & 73.4 & 58.9      & 45.9    & 59.7      \\ \cline{2-6} 
                                   & Baseline$^{\bigstar}$        & 74.4 & 60.2      & 47.6   & 60.7     \\ \cline{2-6} 
      ResNet                        & Ours & 74.5 & 59.5      & 47.4    & 60.8      \\ \cline{2-6} 
       101~\cite{he2016deep}        & Ours$^{\bigstar}$     & 75.5 & 60.7      & 49.0   & 61.7     \\ \cline{2-6} 
                                   & +         & 1.1  & 0.6       & 1.59    & 1.1       \\ \cline{2-6} 
                                   & +$^{\bigstar}$   & 1.1  & 0.5       & 1.4     & 1.0      \\ \hline

                                   & Baseline  & 74.7 & 61.5      & 48.9    & 61.5      \\ \cline{2-6} 
                                   & Baseline$^{\bigstar}$        & 75.4 & 62.6      & 50.3    & 62.2      \\ \cline{2-6} 
  ResNext                       & Ours & 75.8 & 62.8      & 50.2    & 62.6     \\ \cline{2-6} 
   101\_32x8d                         & Ours$^{\bigstar}$       & 76.4 & 63.5      & 51.3    & 63.0      \\ \cline{2-6} 
  ~\cite{xie2017aggregated}     & +         & 1.1  & 1.3       & 1.3     & 1.1       \\ \cline{2-6} 
                                   & +$^{\bigstar}$   & 1.0  & 0.9       & 1.0    & 0.8       \\ \hline
\end{tabular}
}
\end{table}  

\subsection{Experiments on Different Datasets}
\vspace{-0.1cm}
\paragraph{NYUDv2 Dataset.} We adopted two popular settings for this dataset, i.e., 13-class~\cite{silberman2012indoor} and 40-class~\cite{gupta2013perceptual}, and show the results of baseline and our method with different backbones on NYUDv2-13 and NYUDv2-40 in Table~\ref{tab:nyu-13} and Table~\ref{tab:nyu-40}, respectively. 
It can be seen that architectures with ShapeConv outperform the baselines with a large margin under all settings.

We also compare the performance of our ShapeConv with several recently developed methods in Table~\ref{tab:NYUDv2-13 performance} and Table~\ref{tab:NYUDv2-40 performance}. As illustrated in Table~\ref{tab:NYUDv2-13 performance}, ShapeConv achieves the best over all the four metrics on NYUDv2-13. Compared to the recently proposed  method~\cite{zhao2018dense}, our approach yields around 6.3\% improvements on Mean IOU which is the most commonly used metric for semantic segmentation. In addition, our method also achieves a competitive performance on NYUDv2-40 in Table~\ref{tab:NYUDv2-40 performance}. 

\vspace{-0.2cm}
\begin{table}[h!]
 
   \caption{
        \label{tab:NYUDv2-13 performance}
        Performance comparison with other methods on NYUDv2-13 dataset.}
     \centering
  \scalebox{0.85}{
      \begin{tabular}{c | c|c | c|c  }
       \hline 
         \multirow{2}{*}{Method}  & Pixel  & Mean   & Mean  & f.w. \\
                                 & Acc.(\%) &Acc.(\%)  &IoU.(\%)  &IoU.(\%) \\
        \hline
        Eigen~\cite{eigen2015predicting} &75.4 &66.9 &- &- \\
        \hline
        MVCNet~\cite{ma2017multi}       & 77.8    &69.5     &   57.3   &- \\
        \hline 
        Ours  & \textbf{82.6} & \textbf{75.7}      & \textbf{65.1}    & \textbf{71.2}\\
        \hline
        \hline
        MVCNet~\cite{ma2017multi}$^{\bigstar}$       & 79.1    &70.6     &   59.1   &- \\
        \hline 
        PVNet~\cite{zhao2018dense}$^{\bigstar}$  &82.5 &74.4 &59.3 &- \\
        \hline
        Ours$^{\bigstar}$        &\textbf{ 82.9} & \textbf{76.0 }     & \textbf{65.6 }   & \textbf{71.6 }    \\
        \hline
        
      \end{tabular}
  }
 
\end{table} 

\vspace{-0.4cm}
\begin{table}[h!]
 
   \caption{
        \label{tab:NYUDv2-40 performance}
        Performance comparison with other methods on NYUDv2-40 dataset.}
     \centering
  \scalebox{0.85}{
      \begin{tabular}{c | c|c | c|c  }
       \hline 
         \multirow{2}{*}{Method}  & Pixel  & Mean   & Mean  & f.w. \\
                                 & Acc.(\%) &Acc.(\%)  &IoU.(\%)  &IoU.(\%) \\
        \hline
        FCN~\cite{long2015fully}     & 65.4   &  46.1      &  34.0   &49.5\\
        \hline 
        LSD-GF~\cite{cheng2017locality} &71.9 &60.7 &45.9 &59.3\\
        \hline
        D-CNN~\cite{wang2018depth} & -   &61.1  & 48.4 & -\\
        \hline 
        MMAF-Net~\cite{fooladgar2019multi}  &72.2     & 59.2        & 44.8  &-    \\
        \hline
        ACNet~\cite{hu2019acnet}  &- & - &48.3 &- \\
        \hline
        Ours  & \textbf{75.8} & \textbf{62.8}      & \textbf{50.2}    & \textbf{62.6}  \\
        \hline
        \hline
        CFN~\cite{lin2017cascaded}$^{\bigstar}$     &  -   &  -       & 47.7 &- \\
        \hline 
        3DGNN~\cite{qi20173d}$^{\bigstar}$       & -    & 55.7        & 43.1  &- \\
        \hline 
        RDF~\cite{park2017rdfnet}$^{\bigstar}$   &76.0     &62.8   &50.1  &-\\
        \hline 
        M2.5D~\cite{xing2020malleable}$^{\bigstar}$  &\textbf{76.9} &- &50.9 &-  \\
        \hline
        SGNet~\cite{chen2021spatial}$^{\bigstar}$  & 76.8 &63.3 &51.1 & -\\
        \hline
        Ours$^{\bigstar}$       & 76.4 & \textbf{63.5}      & \textbf{51.3}    & \textbf{63.0}\\
        \hline
        
      \end{tabular}
  }
 
\end{table} 

\paragraph{SUN-RGBD Dataset.}
The comparison results between baseline and ours with SUN-RGBD dataset are reported in Table~\ref{tab:sun}. It can be observed that our ShapeConv also produces a positive effect under all settings. We also compared the performance of ours with several recently developed methods in Table~\ref{tab:SUN-RGBD performance}. It is worth noting that the performance of the ShapeConv-enhanced Network with backbone of ResNet-50 in Table~\ref{tab:sun} has already achieved better results than several methods in Table~\ref{tab:SUN-RGBD performance}, such as 3DGNN-101~\cite{qi20173d} and RDF-152~\cite{park2017rdfnet} which take the ResNet-101 and -152 as backbone, respectively.

\begin{table}[]
\centering
\caption{
\label{tab:sun}
Performance comparison with baselines on SUN-RGBD dataset. The architectures adopted in this table is deeplabv3+ with different backbones.}
\scalebox{0.8}{
\begin{tabular}{c|c|c|c|c|c}
\hline
\multirow{2}{*}{Backbone}  &\multirow{2}{*}{Setting} & Pixel  & Mean   & Mean  & f.w. \\
                           &         & Acc.(\%) &Acc.(\%)  &IoU.(\%)  &IoU.(\%) \\
                        \hline
                     & Baseline  & 81.1 & 56.5      & 45.5    & 69.7      \\ \cline{2-6} 
                        & Baseline$^{\bigstar}$         & 81.4 & 57.5      & 46.6    & 70.0      \\ \cline{2-6} 
     ResNet                   & Ours & 81.6 & 56.8      & 46.3    & 70.3      \\ \cline{2-6} 
     50~\cite{he2016deep}     & Ours$^{\bigstar}$       & 81.9 & 57.9      & 47.7   & 70.6     \\ \cline{2-6} 
                        & +         & 0.5  & 0.3       & 0.8    & 0.6       \\ \cline{2-6} 
                        & +$^{\bigstar}$    & 0.5  & 0.4      & 1.1     & 0.6      \\ \hline
                        & Baseline  & 81.6 & 57.8      & 46.9    & 70.4      \\ \cline{2-6} 
                        & Baseline$^{\bigstar}$         & 81.6 & 58.4      & 47.6   & 70.5      \\ \cline{2-6} 
   ResNet                     & Ours & 82.0 & 58.5      & 47.6    & 71.2      \\ \cline{2-6} 
     101~\cite{he2016deep}    & Ours$^{\bigstar}$        & 82.2 & 59.2      & 48.6    & 71.3      \\ \cline{2-6} 
                        & +         & 0.4  & 0.7       & 0.7     & 0.8      \\ \cline{2-6} 
                        & +$^{\bigstar}$    & 0.6  & 0.8       & 1.0    & 0.8       \\ \hline
   
\end{tabular}
}

\end{table}  

\begin{table}[h!]
\centering

\vspace{-0.2cm}
\caption{
\label{tab:SUN-RGBD performance}
Performance comparison on SUN-RGBD dataset.}
\scalebox{0.85}{
\begin{tabular}{c | c|c | c|c  }
       \hline 
         \multirow{2}{*}{Method}  & Pixel  & Mean   & Mean  & f.w. \\
                                 & Acc.(\%) &Acc.(\%)  &IoU.(\%)  &IoU.(\%) \\
        \hline
        3DGNN-101~\cite{qi20173d}       & -    & 55.7        & 44.1   &-\\
        \hline 
        D-CNN-50~\cite{wang2018depth} & -   &53.5  & 42.0 & -\\
        \hline 
        MMAF-Net-152~\cite{fooladgar2019multi}  &81.0     & 58.2        & 47.0  &-    \\
        \hline
        SGNet-101~\cite{chen2021spatial}  & 81.0 &\textbf{59.8} &47.5 & -\\
        \hline
        Ours-101  & \textbf{82.0}    & 58.5     & \textbf{47.6}  & \textbf{71.2} \\
        \hline
        \hline
        CFN-101~\cite{lin2017cascaded}$^{\bigstar}$     &  -   &  -       & 48.1 &- \\
        \hline 
        3DGNN-101~\cite{qi20173d}$^{\bigstar}$       & -    & 57.0        & 45.9  &- \\
        \hline 
        RDF-152~\cite{park2017rdfnet}$^{\bigstar}$   &81.5     &60.1   &47.7  &-\\
        \hline 
        SGNet-101~\cite{chen2021spatial}$^{\bigstar}$  & 82.0 &\textbf{60.7} &48.6 & -\\
        \hline
        Ours-101$^{\bigstar}$  & \textbf{82.2}    &59.2      &\textbf{48.6}   &\textbf{71.3}  \\
        \hline
        
      \end{tabular}
}
\end{table} 

\vspace{-0.7cm}
\paragraph{SID Dataset.}
Note that SID dataset is much larger than the other two datasets, contributing to a better testbed for evaluating RGB-D semantic segmentation model capabilities. The results on SID dataset between the baseline with ours and the state-of-the-art methods are reported in Table~\ref{tab:SID-performance}. We can observe that our ShapeConv surpasses these methods with a large margin. Note that even though we utilized a strong baseline (ResNet-101 backbone) which surpasses MMAF-Net-152 (ResNet-152 backbone) with 1.7\% Mean IoU, our ShapeConv can still achieves a 6\% Mean IoU improvement. This highlights the effectiveness of our method.

\begin{table}[h!]
  \centering
\caption{
\label{tab:SID-performance}
Performance comparison on SID dataset. The architectures of baseline and ours adopted in this table is deeplabv3+ with ResNet-101 backbone and the ``+'' denote the deltas relative to the baseline method.}
  \scalebox{0.85}{
      \begin{tabular}{c | c | c | c | c }
       \hline 
        \multirow{2}{*}{Method}  & Pixel  & Mean   & Mean  & f.w. \\
                                 & Acc.(\%) &Acc.(\%)  &IoU.(\%)  &IoU.(\%) \\
        \hline 
        D-CNN~\cite{wang2018depth}  & 65.4  &55.5 & 39.5  &49.9\\
        \hline 
        MMAF-Net-152~\cite{fooladgar2019multi}  &76.5     & 62.3        & 52.9  &-    \\
        \hline
        \hline
        Baseline-101  &78.7     &63.2   & 54.6        & 65.6   \\ \hline
        Ours-101  &\textbf{82.7} &\textbf{70.0} &\textbf{60.6} &\textbf{71.2}\\
        \hline 
        +         &  4.0   &    6.8        &   6.0      &   5.6 \\ \hline
      \end{tabular}
  }
  
\end{table}  

\subsection{Experiments on Different Architectures}
\label{sec:diff ar}
\vspace{-0.2cm}
Our proposed ShapeConv is a general layer for RGB-D semantic segmentation which can be easily plugged into most CNNs as a replacement for the vanilla convolution in semantic segmentation. To verify its generalization properties, we also evaluated the effectiveness of our method in several representative semantic segmentation architectures: Deeplabv3+~\cite{chen2018encoder}, Deeplabv3~\cite{chen2017rethinking}, UNet~\cite{ronneberger2015u}, PSPNet~\cite{zhao2017pyramid} and FPN~\cite{lin2017feature} with different backbones (ResNet-50~\cite{he2016deep}, ResNet-101~\cite{he2016deep}) on NYUDv2-40 dataset, and reported the performance in Table~\ref{tab:baseline-backbone}. We can see that ShapeConv brings significant performance improvements under all settings, demonstrating the generalization capability of our method.

\begin{figure*}[t!]
	\centering
	\centering
	\includegraphics[width=\textwidth]{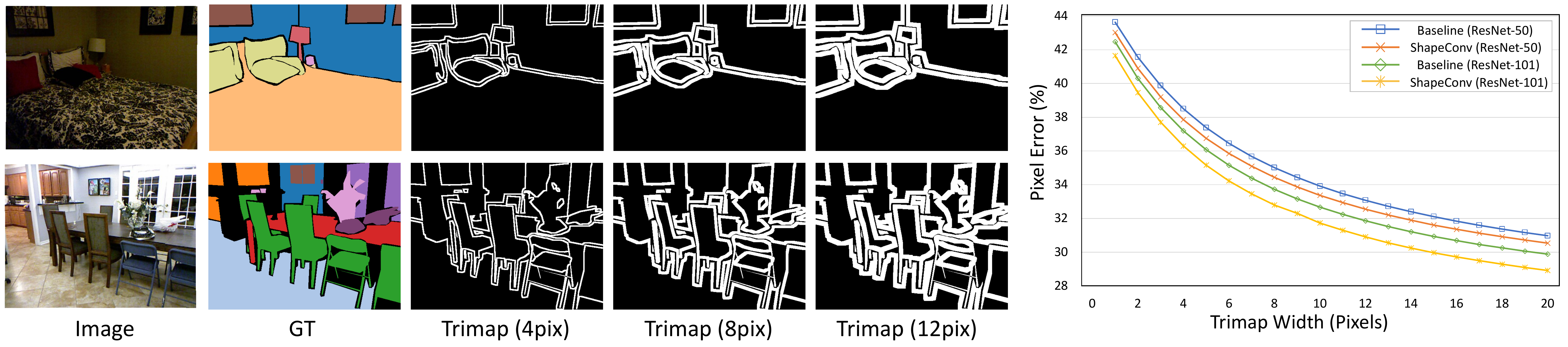}
	\vspace{-0.035\linewidth}
	\caption{Segmentation accuracy around object boundaries. In this figure, the left is the visualization of the ``trimap'' measure; The right is the percent of misclassified pixels within trimaps of different widths.}
	\label{fig:trimap}
\end{figure*}

\begin{table}[h!]
\centering
\caption{
\label{tab:baseline-backbone}
Performance comparison with different baseline methods on NYUDv2-40 dataset.}
\scalebox{0.70}{
\begin{tabular}{c|c|c|c|c|c|c}
\hline
\multirow{2}{*}{Architecture} &Back &\multirow{2}{*}{Setting} & Pixel  & Mean   & Mean  & f.w. \\
                  &  bone    &         & Acc.(\%) &Acc.(\%)  &IoU.(\%)  &IoU.(\%) \\
\hline
                    &Res  & Baseline      & 73.4 & 58.9      & 45.9    & 59.7      \\ \cline{3-7} 
                    &Net  & Ours     & 74.5 & 59.5     & 47.4   & 60.8      \\ \cline{3-7} 
 Deeplabv3+         & 101 & +             & 1.1  & 0.6       & 1.5     & 1.1       \\ \cline{2-7} 
 ~\cite{chen2018encoder} &Res   & Baseline      & 73.1 & 57.7      & 45.6   & 59.2     \\ \cline{3-7} 
                         &Net   & Ours     & 74.1 & 59.1     & 47.3  & 60.5     \\ \cline{3-7} 
                         &50    & +            & 1.0  & 1.4     & 1.7   & 1.3       \\ \hline
                         &Res   & Baseline      & 73.3 & 57.3      & 45.1    & 59.2     \\ \cline{3-7} 
                         &Net   & Ours     & 73.6 & 58.5      & 46.4    & 59.7      \\ \cline{3-7} 
 Deeplabv3               &101   & +             & 0.3 & 1.2       & 1.3    & 0.5       \\ \cline{2-7} 
 ~\cite{chen2017rethinking} & Res  & Baseline      & 71.6 & 55.5      & 43.2    & 57.2      \\ \cline{3-7} 
                            &Net   & Ours     & 72.8 & 56.6     & 44.9    & 58.5     \\ \cline{3-7} 
                            & 50   & +       & 1.2  & 1.1      & 1.7    & 1.3       \\ \hline
                          &Res & Baseline    & 70.9 & 54.7      & 42.1    & 57.7     \\ \cline{3-7} 
                          &Net & Ours   & 72.3 & 56.5     & 43.9   & 58.8      \\ \cline{3-7} 
UNet                      & 101     & +      & 1.4  & 1.8        & 1.8     & 1.1      \\ \cline{2-7} 
 ~\cite{ronneberger2015u} &Res  & Baseline      & 70.0 & 51.7      & 39.7   & 55.5     \\ \cline{3-7} 
                          &Net  & Ours     & 70.8 & 54.1     & 42.0    & 56.9      \\ \cline{3-7} 
                          &50   & +      & 0.8 & 2.4        & 2.3     & 1.4      \\ \hline
                          &Res   & Baseline      & 72.8 & 56.8      & 44.2    & 58.9      \\ \cline{3-7} 
                          &Net  & Ours     & 73.3 & 59.2      & 46.3    & 59.6      \\ \cline{3-7} 
PSPNet                    &101  & +             & 0.5  & 2.4      & 2.1    & 0.7     \\ \cline{2-7} 
 ~\cite{zhao2017pyramid}  &Res  & Baseline      & 71.1 & 53.6      & 42.0    & 56.7      \\ \cline{3-7} 
                          &Net  & Ours     & 72.0 & 56.2      & 44.0    & 57.7     \\ \cline{3-7} 
                          &50   & +             & 0.9 & 2.6       & 2.0      & 1.0      \\ \hline
                          &Res  & Baseline & 72.8 & 57.3      & 44.7    & 59.1     \\ \cline{3-7} 
                          &Net  & Ours     & 73.6 & 58.4      & 45.9   & 60.0     \\ \cline{3-7} 
 FPN                      & 101 & +             & 0.8  & 1.1      & 1.2   & 0.9      \\ \cline{2-7} 
 ~\cite{lin2017feature}   & Res & Baseline      & 70.3 & 52.8      & 40.9    & 56.0      \\ \cline{3-7} 
                          & Net & Ours     & 71.5 & 54.9     & 42.8    & 57.5     \\ \cline{3-7} 
                          & 50  & +             & 1.2  & 2.1      & 1.9    & 1.5      \\ \hline
\end{tabular}
}
\end{table}  

\subsection{Visualization}
\vspace{-0.2cm}
Figure~\ref{fig:vis-seg} illustrates the qualitative results on NYUDv2-13 and -40, more results can be found in the Supp. As shown in this figure, the depth information, especially the detailed one, can be well utilized by ShapeConv to extract the object features. For instance, the chair and table regions in the top example of Figure~\ref{fig:vis-seg}(a) are with gradually changed colors, making it hard to predict accurate segmentation boundaries of the baseline method. The shape features learned by ShapeConv makes the accurate cut following the geometric hints compare with the conventional convolutional layer. For other two cases, i.e., the chair in the bottom example of Figure~\ref{fig:vis-seg}(a) and the desk in the top example of Figure~\ref{fig:vis-seg}(b), the ShapeConv can also significantly improve the segmentation results in edge areas compared with the baseline. It is worth noting that for the multiple bookshelves in the bottom example of Figure~\ref{fig:vis-seg}(b), ShapeConv achieves more consistent predictions. This is because our ShapeConv yields a positive tendency for smoothing neighborhood regions within same classes.

To validate the effectiveness of our method on modeling the depth information, we adopted the comparison strategy proposed by Kohli \etal~\cite{kohli2009robust}. Specifically, we counted the relative number of misclassified pixels within a narrow band (``trimap'') surrounding ground-truth object boundaries. As shown in Figure~\ref{fig:trimap}, our method outperforms the baseline across all trimap widths. This further demonstrates the segmentation effectiveness of our method on edge areas, where the shape information matters.

\vspace{-0.1cm}
\subsection{Ablation Study}
\vspace{-0.2cm}
We conducted ablation experiments to validate the indispensability of the two introduced weights in Equation~\ref{eq:shapeconv-kernel}. As can be observed in Table~\ref{tab:wb-ws-ab}, the model performance degrades when removing either $\mathbb{W}_B$ or $\mathbb{W}_S$, or both. This proves that both the base-kernel and shape-kernel are essential for the final performance improvement, and combing these two achieves the best results.

\vspace{-0.2cm}
\begin{table}[h]
	\centering
		\caption{\label{tab:wb-ws-ab}
	Performance comparison with and without $\mathbb{W}_B$ and $\mathbb{W}_S$ in ShapeConv on NYUDv2-40. The architecture adopted in this table is deeplabv3+ with ResNet-101 as backbone.}
	\scalebox{0.8}{
	\begin{tabular}{ c c | c c c c}
	\hline 
        \multirow{2}{*}{$\mathbb{W}_B$} &\multirow{2}{*}{$\mathbb{W}_S$}  & Pixel  & Mean   & Mean  & f.w. \\
        &      & Acc.(\%) &Acc.(\%)  &IoU.(\%)  &IoU.(\%) \\
        \hline
         &  & 73.4 & 58.9      & 45.9    & 59.7      \\ 
       \hline
        $\checkmark$ & & 73.9 & 59.4 &47.0  &60.1   \\
        \hline
        &$\checkmark$ &74.1 &59.2  &46.3 & 60.1  \\
        \hline
        $\checkmark$ &$\checkmark$ & 74.5 & 59.5      & 47.4    & 60.8  \\
        \hline
	\end{tabular}
	}
\end{table}
\vspace{-0.4cm}

To provide a more in-depth analysis of ShapeConv, we conducted detailed ablation studies on the NYUDv2-40 dataset with deeplabv3+ and ResNet-101 as baseline and backbone, respectively. Results on more datasets can be found at the Supp. Table~\ref{tab:ab-ex-nyu-40} illustrates the results and the key observations from this table are as follows: 1) The input features with HHA outperform the Depth images for the baseline and ours; 2) Replacing the vanilla convolution with ShapeConv leads to considerable performance improvements on both Depth and HHA; 3) The multi-scale setting in testing phase brings more performance gains; 4) Cascading the ShapeConv with HHA and multi-scale testing can achieve the best result.

\begin{table}[]
\centering
\caption{
\label{tab:ab-ex-nyu-40}
Ablation study of the proposed ShapeConv on the NYUDv2-40 dataset. RGB, Detph and HHA denote the inputs consisting of RGB images, depth images and HHA images.}
\scalebox{0.75}{
\begin{tabular}{l|l|l|l|l}
\hline
\multirow{2}{*}{Setting}  & Pixel  & Mean   & Mean  & f.w. \\
                                 & Acc.(\%) &Acc.(\%)  &IoU.(\%)  &IoU.(\%) \\
\hline
$\mathcal{a}$.RGB                                 & 71.8 & 56.9     & 43.9    & 57.3                                                    \\ 
\hline
$\mathcal{b}$.RGB+Depth                       & 72.8 &58.9 &44.9&57.7 \\ 
\hline
$\mathcal{c}$.RGB+Depth$^{\bigstar}$ &  73.9 &59.1 &46.8 &60.0\\
\hline

$\mathcal{d}$.RGB+HHA                             & 73.4 & 58.9     & 45.9    & 59.7                                                   \\ 
\hline
$\mathcal{e}$.RGB+HHA$^{\bigstar}$           & 74.4 & 60.2     & 47.6    & 60.7                                                    \\ 
\hline
$\mathcal{f}$.RGB+Depth+ShapeConv             &73.9 &58.2 &46.2 &60.0 \\ 
\hline
$\mathcal{g}$.RGB+Depth+ShapeConv$^{\bigstar}$   &74.8 &59.2 &47.5 &60.8\\ 
\hline
$\mathcal{h}$.RGB+HHA+ShapeConv                   & 74.5 & 59.5      & 47.4    & 60.8                                                   \\ 
\hline
$\mathcal{i}$.RGB+HHA+ShapeConv$^{\bigstar}$                 & 75.5 & 60.7      & 49.0    & 61.7                                                   \\ 
\hline
\end{tabular}
}
\end{table}  

%% file: 06_con.tex
\section{Conclusion}
\vspace{-0.2cm}
In this paper, we propose a ShapeConv layer to effectively leverage the depth information for RGB-D semantic segmentation. In particular, an input patch is firstly decomposed into two components, i.e., \emph{shape} and \emph{base}, which are then decorated with two corresponding learnable weights before the convolution is applied. We have conducted extensive experiments on several challenging indoor RGB-D semantic segmentation benchmarks and promising experimental results can be observed. Moreover, it is worth noting that our ShapeConv introducing no additional computation or memory in comparison with the vanilla convolution during inference, yet with superior performance.

In fact, the shape-component is inherent in the local geometry and highly relevant to the semantics in images. In the future, we plan to expand the application scope to other geometry entities, such as point clouds, where the shape-base decomposition is more challenging due to the additional degree of freedom.